\documentclass[11pt,a4paper]{article}
\usepackage{times,latexsym}
\usepackage{url}
\usepackage[T1]{fontenc}

\usepackage{comment}
\usepackage[ruled,vlined]{algorithm2e}

\usepackage{multirow}
\usepackage{pgf}
\usepackage{tikz}

\usepackage{mathabx}
\usepackage{graphicx}
\usepackage{tabularx}

\usepackage[acceptedWithA]{tacl2018v2}

\usepackage{xspace,mfirstuc,tabulary}

\definecolor{teal}{rgb}{0.0, 0.5, 0.5}

\newcommand{\modelName}{ProoFVer}

\newif\iftaclinstructions
\taclinstructionsfalse 
\iftaclinstructions
\renewcommand{\confidential}{}
\renewcommand{\anonsubtext}{(No author info supplied here, for consistency with
TACL-submission anonymization requirements)}
\newcommand{\instr}
\fi

\iftaclpubformat 

\else

\fi

\title{\modelName: Natural Logic Theorem Proving for Fact Verification}

\author{
 Amrith Krishna
 \\
 Department of Computer Science, \\
 University of Cambridge \\
  \texttt{ ak2329@cam.ac.uk} \\\\\And
  Sebastian Riedel\\
  Meta AI and \\
  University College London \\
  \texttt{sriedel@fb.com}\\\And
  Andreas Vlachos \\
  Department of Computer Science, \\
  University of Cambridge \\
  \texttt{av308@cam.ac.uk}
}

\date{}

\begin{document}
\maketitle
\begin{abstract}

Fact verification systems typically rely on 
neural network classifiers for veracity prediction which
lack explainability.
This paper proposes \modelName, which uses a
seq2seq model to generate natural logic-based inferences as proofs.
These proofs consist of 
lexical mutations between spans in the claim and the evidence retrieved, each  marked with 
a natural logic operator. Claim veracity is determined solely based on the sequence of these operators. Hence, these proofs are faithful explanations,
and this makes \modelName~faithful by construction.
Currently, \modelName\ has the highest label accuracy and the second best Score in the FEVER leaderboard. 
Furthermore, it improves by 13.21\% points over the next best model on a dataset with counterfactual instances, demonstrating its robustness. 
As explanations, the proofs 
show better overlap with human rationales than attention-based highlights and the proofs help humans predict model decisions correctly more often than
using the
evidence directly.\footnote{Find our code and data at \url{https://github.com/krishnamrith12/ProoFVer}.}

\end{abstract}

\section{Introduction}

Fact verification systems 
typically comprise an evidence retrieval model followed by a textual entailment classifier~\cite{thorne-etal-2018-fact}. Recent high performing fact verification systems~\cite{zhong-etal-2020-reasoning,ye-etal-2020-coreferential} use 
neural models for textual entailment whose reasoning 
is opaque to humans despite advances in interpretablity~\cite{han-etal-2020-explaining}. On the other hand, proof systems like  NaturalLI~\cite{angeli-manning-2014-naturalli}  provide transparency in their decision making
for entailment tasks, 
by using explicit proofs in the form of 
natural logic. However, the accuracy of such approaches
often does not match 
that of  neural models~\cite{abzianidze-2017-langpro}.




Justifying  decisions is central to fact verification~\cite{uscinski2013epistemoology}.  While models such as  those developed for FEVER~\cite{thorne-etal-2018-fact} typically substantiate their decisions by presenting the evidence as is, more recent proposals use the evidence to  generate explanations. Here, models highlight salient parts of the evidence~\cite{popat-etal-2018-declare, wu-etal-2020-dtca}, generate summaries~\cite{kotonya-toni-2020-explainable,atanasova-etal-2020-generating-fact}, correct factual errors~\cite{thorne2021factual,schuster-etal-2021-get}, answer claim related questions ~\cite{fan-etal-2020-generating}, or perform rule discovery~\cite{DBLP:conf/tto/AhmadiLPS19,10.1145/3289600.3290996/ExFakt}. An explanation is faithful only if it reflects the information  that is used for decision making~\cite{lipton2018faithful,jacovi-goldberg-2020-towards}, which these systems do not guarantee. A possible exception here would be the rule discovery models
, however, their performance often suffers due to limited knowledge base coverage and/or the noise in rule extraction from text~\cite{kotonya-toni-2020-explainable-automated,pezeshkpour2020revisiting}. 
Faithful explanations 
are useful as mechanisms to dispute, debug or advice~\cite{jakoviGoldberg10.1162/tacl_a_00367}, which may aid a news agency for advice, a user to dispute decisions, and a developer for model debugging in fact verification. 


\begin{figure*}
    \centering
    \includegraphics[width=0.9\textwidth,page=6,trim={0 5cm 2cm 0}]{images/figures.pdf}
    \caption{
    The proof generator in \modelName, generates the natural logic proofs using a seq2seq model. The natural logic operators from the proof are used as transitions in the DFA to determine the veracity of the claim. The states {\sc S}, {\sc N}, and {\sc R} in the automaton denote the task labels {\sc Supports}, {\sc Refutes}, and {\sc Not Enough Info} respectively. The transitions in the automaton are the natural logic operators (NatOPs) defined in Table \ref{tab:natlogic}.
   }
    \label{fig:working}
\end{figure*}

\begin{figure}[!ht]
    \centering
    \includegraphics[width=0.9\textwidth,page=12,trim={0 5.2cm 2cm 0.3cm}]{images/figures.pdf}
    \caption{Proof steps for the input in Figure \ref{fig:working}.} 
    \label{fig:proof}
\end{figure}

 Keeping both accuracy and explainability in mind, 
we propose  \textbf{\modelName} - \textbf{Proo}f System for \textbf{F}act \textbf{Ver}ification 
which generates proofs  or refutations of the claim given evidence as natural logic based inference. 
\modelName~
follows the natural logic based theory of compositional entailment, originally proposed in NatLog~\cite{maccartney-manning-2007-natural}. 
 In the example of Figure~\ref{fig:working} 
 \modelName~generates the proof shown in Figure~\ref{fig:proof}, for a given claim and evidence. Here, at each step in the proof, a claim span is mutated  with a span from the evidence. Each such mutation is marked with an entailment relation, by assigning a natural logic operator~\cite[NatOp,][]{angeli-manning-2014-naturalli}. A step in the proof can be represented using a triple, consisting of the aligned spans in the mutation and its assigned NatOp. 
In the example, the mutations in the first and last triples occur with semantically equivalent spans, 
and hence are assigned with the equivalence NatOp~($\equiv$). However, the mutation in the second triple results in a contradiction, as `short story' is replaced with `novel' and an item  cannot be both. Hence, the mutation is assigned the alternation NatOp~($\updownharpoons$). The sequence of NatOps from the proof become the transitions in the DFA shown in Figure~\ref{fig:working}, which in this case terminates at the `{\sc Refute~(R)}' state, i.e.\ the evidence refutes the claim.

Unlike other natural logic systems~\cite{angeli-etal-2016-combining,feng-etal-2020-exploring}, \modelName~can form a proof by combining spans from multiple evidence sentences, by leveraging the entity mentions linking those sentences.
The proof is generated by a seq2seq model trained using a heuristically annotated dataset, obtained by combining 
information from the publicly available FEVER dataset~\citep{thorne-etal-2018-fever,thorne2021factual}
with PPDB~\cite{pavlick-etal-2015-ppdb2}, Wordnet~\cite{10.1145/219717.219748wordnet} and Wikidata~\cite{10.1145/2629489Wikidata}. We heuristically generate 
the training data for the claims in three datasets, namely, FEVER, symmetric FEVER~\cite{schuster-etal-2019-towards}, and FEVER 2.0~\cite{thorne-etal-2019-fever2}.


\modelName~currently is the highest scoring system on the FEVER leaderboard in terms of label accuracy and is the second best system in terms of FEVER score. Additionally, \modelName~has robustness and explainability as its key strengths. Its veracity predictions are solely  determined using the generated proof. Hence by design, \modelName's proofs, when used as explanations, are faithful by construction~\cite{lei-etal-2016-rationalizing, jain-etal-2020-learning}.  Similarly, it demonstrates robustness to counterfactual instances from Symmetric FEVER and adversarial instances from FEVER 2.0. In particular, \modelName~achieved 13.21\% higher label accuracy than that of the next best model~\cite{ye-etal-2020-coreferential} for symmetric FEVER and similarly improves upon the previous best results~\cite{schuster-etal-2021-get} on Adversarial FEVER.

To evaluate the robustness of fact verification systems against the impact of superfluous information from the retriever, we propose a new metric, Stability Error Rate (SER), which measures the proportion 
of instances where superfluous information changes the decision of the model. 
\modelName~achieves a SER of 5.73\%, compared to 9.36\% of \newcite{stammbach-2021-evidence}, where a lower SER is preferred. \modelName's proofs as explanations, apart from being faithful, score high in their overlap with human rationales with a token overlap F1-Score of 93.28\%, 5.67\% points more than attention-based highlights from \newcite{ye-etal-2020-coreferential}. Finally, 
humans, with no knowledge of natural logic, correctly predict \modelName's decisions  81.67\% of the times compared to 69.44\% when  using the retrieved evidence.

 \begin{figure*}
    \centering
    \includegraphics[width=0.9\textwidth,page=9,trim={0 4.4cm 1cm 0cm}]{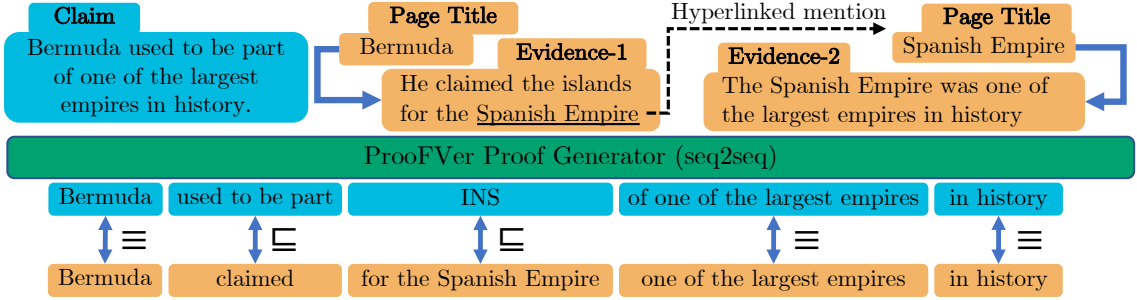}
    \caption{A claim requiring multiple evidence sentences for verification.} 
    \label{fig:multiple}
\end{figure*}

\section{Natural Logic Proofs as Explanations}
\label{NatlogSection}

 %
 
%
 Natural logic operates directly on natural language~\cite{angeli-manning-2014-naturalli,abzianidzelasha}. Thus it 
 is appealing for fact verification, as structured knowledge bases like Wikidata typically lag behind text-based encyclopedias such as Wikipedia in terms of coverage~\cite{johnson2020analyzing}.  Furthermore, it obviates the need to translate claims and evidence into meaning representations such as lambda calculus~\cite{10.5555/3020336.3020416lambdazettlem}. While such representations may be 
 more expressive, they require the development of semantic parsers, introducing another source of potential errors
 in the verification process.

Natural Logic 
has been previously employed in several information extraction and NLU tasks such as Natural Language Inference~\cite[NLI,][]{abzianidze-2017-langpro,feng-etal-2020-exploring}, question answering~\cite{angeli-etal-2016-combining} and open information extraction~\cite[Chapter 5]{angeli-thesis}. 
NatLog~\cite{maccartney-manning-2007-natural}, building on earlier theoretical work on natural logic and monotonicity calculus~\cite{van1986essays,valencia1991studies},   uses natural logic for 
textual inference. 

NaturalLI~\cite{angeli-manning-2014-naturalli} extended NatLog by adopting the formal semantics of \newcite{icard-iii-moss-2014-recent}, and it is a proof system formulated for the NLI task. It determines  the entailment of a hypothesis by searching over a database of 
premises. The proofs are in the form of a natural logic based logical inference, which results in a sequence of mutations between a premise and a hypothesis. Each mutation is marked with a natural logic relation, and is realised as a lexical substitution, 
forming a step in the inference. Each mutation results in a new sentence, and the natural logic relation assigned to it identifies the type of entailment that holds between the sentences before and after the mutation. NaturalLI adopts a set of seven natural logic operators, as shown in Table \ref{tab:natlogic}. 
The operators were originally proposed in NatLog~\cite[p. 79]{maccartney2009natural}. We henceforth refer to these operators as \textit{NatOps}. 
\begin{table}[!ht]

\begin{tabular}{|l|c|}
\hline
NatOP: Name                       & Definition                                                                            \\ \hline
$\updownharpoons$: Alternation    & \begin{tabular}[c]{@{}c@{}}$x \! \cap \! y \! = \! \emptyset \! \wedge \! x \! \cup \! y \! \neq \! U$\end{tabular} \\ \hline
$\smile$: Cover                   & \begin{tabular}[c]{@{}c@{}}$x \! \cap \! y \! \neq \! \emptyset \! \wedge \! x \! \cup \! y \! = \! U$\end{tabular} \\ \hline
$\equiv$: Equivalence             & $x = y$                                                                               \\ \hline
$\sqsubseteq$: Forward Entailment & $x \subset y$                                                                         \\ \hline
$\curlywedge$: Negation           & \begin{tabular}[c]{@{}c@{}}$x \! \cap \! y \! = \! \emptyset \! \wedge \! x \! \cup \! y \! = \! U$\end{tabular}    \\ \hline
$\sqsupseteq$: Reverse Entailment & $x \supset y$                                                                         \\ \hline
$\hash$: Independence             & All other cases                                                                       \\ \hline
\end{tabular}
\caption{ Natural logic relations (NatOps) and their set
theoretic definitions.}
\label{tab:natlogic}
\end{table}

 To determine whether a hypothesis is entailed by a premise, NaturalLI  uses a deterministic finite state automaton (DFA). Here, each state is an entailment label,
and the transitions are the NatOps (Figure \ref{fig:working}). 
The sequence of NatOps in the inference
is used to traverse the DFA, and the state where it terminates decides the label of the hypothesis-premise pair. 
The decision making process  relies solely on the steps in the logical inference, and thus 
form faithful explanations. 

Other proof systems that apply 
mutations between text sequences have been previously explored. \citet{stern-etal-2012-efficient} explored how to transform a premise into a hypothesis using mutations, however their approach was limited to two-way entailment instead of three-way that is handled by NaturalLI.
Similar proof systems have used mutations in the form of tree-edit operations~\cite{mehdad-2009-automatic}, transformations over syntactic parses~\cite{heilman-smith-2010-tree,harmeling_2009}, knowledge-based transformations in the form of  lexical mutations, entailment rules, rewrite rules, or their combinations ~\cite{bar-haim-etal-2007-semantic,szpektor-etal-2004-scaling}.

\section{\modelName}
\label{modelDesc}


\modelName\ 
uses a seq2seq generator that generates a proof in the form of natural-logic based logical inference, which becomes the input to a deterministic finite state automaton (DFA) for predicting the veracity
of the claim. 
We elaborate on the proof generation process in Section~\ref{ssec:explanation_generation}, and on the veracity prediction 
in Section~\ref{dfaDecision}.



\subsection{Proof Generation}
\label{ssec:explanation_generation}

The proof generator, as shown in Figures~\ref{fig:working} and~\ref{fig:multiple},  takes as input a claim along with one or more retrieved evidence sentences. 
It generates the steps of the proof as a sequence of triples, each 
consisting of a span from the claim, a span from the evidence and a NatOp.  The claim span being substituted and the evidence span replacing it form a \textit{mutation}, and each mutation is assigned a NatOp.
In a proof, we start with the claim, and the mutations are iteratively applied from left to right. 
 Figure~\ref{fig:working} shows a proof containing a sequence of three triples. The corresponding mutated statements 
 at each step of the proof, along with the assigned NatOps, are shown in Figure \ref{fig:proof}. 

We use a seq2seq model following an autoregressive formulation for the proof generation. 
In the proof, successive spans of the claim form part of the successive triples. However, the corresponding evidence spans in the successive triples need not follow any order. As shown in Figure~\ref{fig:multiple}, the evidence spans may come from multiple sentences, and may not all end up being used. 
Finally, the NatOps, as shown in Table \ref{tab:natlogic}, are represented using a predetermined set of tokens.

To get valid proofs during prediction, we need to lexically constrain the inference process 
by switching between three different search spaces depending on which element of the triple is being predicted. To achieve this, we employ dynamically constrained markup decoding \cite{cao2021autoregressiveGENRE}, a modified form of lexically constrained decoding~\cite{post-vilar-2018-fast}. 
This decoding uses markups to switch between the search spaces, and we use the delimiters ``\{'', ``\}'', ``['', and ``]'' as the markups. Using these markups,  we constrain the tokens predicted between a ``\{'' and ``\}'' to be from the claim, between a ``['', and ``]'' to be from the evidence, and the token after ``]'' to be a NatOp token. 
The prediction of a triple begins with predicting a ``\{'', and it proceeds 
by generating a claim span where the tokens are monotonically copied from the claim in the input, until a ``\}'' is predicted. The prediction then continues by generating a ``['' which initiates the evidence span prediction in the triple. The evidence span can begin with any word from the evidence, and is then expanded by predicting subsequent tokens, until 
``]'' is predicted. 
Finally, the NatOp token is predicted. In the next triple, copying resumes from the next token in the claim.  All triples  until the one with the last token in the claim are generated in this manner.
\subsection{Veracity Prediction}
\label{dfaDecision}

The DFA shown in Figure~\ref{fig:working}
uses the  sequence of NatOps predicted by the proof generator as transitions to arrive at the outcome. Figure~\ref{fig:proof} shows the corresponding sequence of transitions for the claim and evidence from   Figure~\ref{fig:working}. Based on this, the DFA in Figure~\ref{fig:working} determines that  the evidence refutes the claim, i.e.\ it terminates in state $R$. NaturalLI~\cite{angeli-manning-2014-naturalli} designed the DFA for the three classes in the NLI classification task, namely entail, contradict and neutral. Here, 
we replace them with {\sc Support (S)}, {\sc Refute (R)}, and {\sc Not Enough Info (N)} respectively  for fact verification. 
\citet{angeli-manning-2014-naturalli} 
chose not to distinguish between negation ($\curlywedge$) and alternation ($\updownharpoons$) relations for NLI,
and assign  $\updownharpoons$ for both. However, there is a clear distinction between cases where each of these NatOPs is applicable in fact verification, and thus we treat them as different NatOps. For instance, in the second mutation for the claim in Figure~\ref{fig:working}, an evidence span ``is not a short story'' would be assigned negation ($\curlywedge$), and not the  currently assigned alternation ($\updownharpoons$) for the mutation with the evidence span ``is a novel''. However, we follow \citet{angeli-manning-2014-naturalli} in not using the cover~($\smile$) NatOp. In rare occasions where this NatOp would be applicable, 
say in a mutation with the spans ``not a novel'' and ``fiction'', we currently assign the independence NatOp ($\hash$).




\section{Generating Proofs for Training} 
\label{sec:explanation_training}

 \begin{figure*}[t]
    \centering
    \includegraphics[width=0.9\textwidth,page=10,trim={0 2cm 1cm 0cm}]{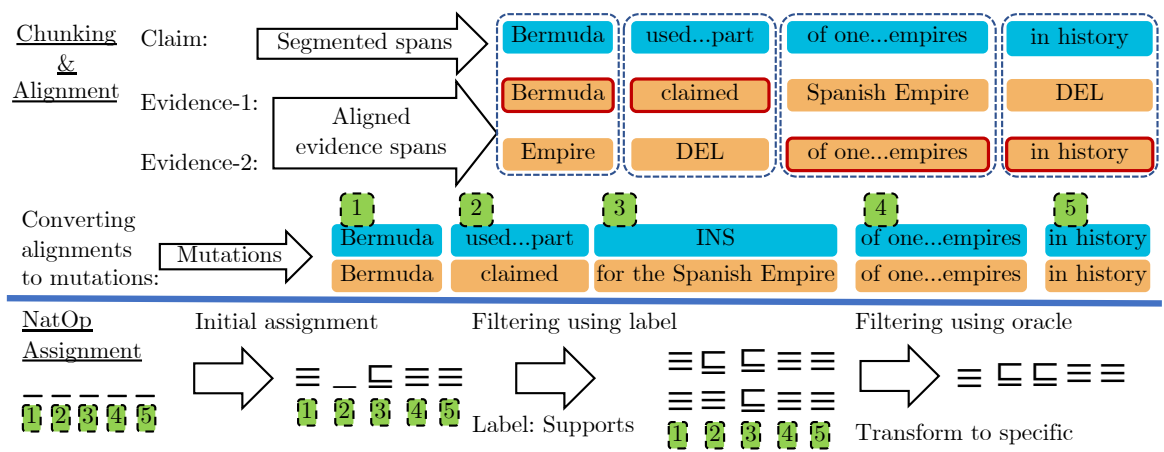}
    \caption{Annotation process for obtaining the proof for the input in Figure \ref{fig:multiple}. It proceeds in two steps, chunking \& alignment, and NatOp assignment, and the latter proceeds by initial mutation assignment and two filtering steps. }
    \label{fig:annotation}
\end{figure*}

Training datasets for evidence-based fact verification consist of instances containing a claim, a label indicating its veracity, and the evidence, typically a set of sentences~\cite{thorne-etal-2018-fever,hanselowski-etal-2019-richly,wadden-etal-2020-fact}. However, we need sequences of triples to train the proof generator of Section~\ref{ssec:explanation_generation}. Manually annotating them would be laborious; thus, we heuristically generate them from existing resources. 
As shown in Figure~\ref{fig:annotation}, we perform a two-step annotation process:  chunking and alignment, followed by the  NatOp assignment. 

\subsection{Chunking and Alignment}
\label{ssec:chunkALign}
 

Chunking the claim into spans is conducted 
using the chunker of~\citet{akbik2019flair}, and any span that does not contain any 
content words is merged with its subsequent span. 
Next, as shown in Figure~\ref{fig:annotation}, a word aligner  ~\cite{jalili-sabet-etal-2020-simalign} aligns each evidence sentence in the input separately with the claim. For each claim span, each evidence sentence provides an aligned  span by grouping together words  that are aligned to it, including any words in between to ensure contiguity.  However, if the aggregated similarity score 
from the aligner for a given pair of claim and evidence spans  falls below an empirically set threshold, then it is ignored and instead the claim span is aligned with the string ``{\sc Del}''.  In Figure \ref{fig:annotation}, ``{\sc Del}'' appears once in each of the evidence sentences.








Next, we convert the alignments into a sequence of mutations, which requires no additional effort in instances with only one evidence sentence. However, a claim span may have multiple evidence spans aligned with it in cases with multiple evidence sentences, as shown in Figure~\ref{fig:annotation}. Here, for a claim span,  we generally select the evidence span with the highest cosine similarity with it. Such spans are marked with solid red borders in Figure~\ref{fig:annotation}. Further, we assume that the evidence sentences 
are linked via entity mentions, such as ``Spanish Empire'' the only hyperlinked mention (from Evidence-1 to 2) in Figure~\ref{fig:multiple}. These hyperlinked mentions must always be added as a mutation, as they provide the context for switching the source of the evidence from one sentence to another. In Figure~\ref{fig:multiple}, ``Spanish Empire'' is not selected as an alignment based on the similarity scores with the claim spans. Hence, it is inserted as the third mutation, at the juncture at which the switch from Evidence-1 to 2 happens. It is aligned with  the string ``{\sc Ins}' in the place of a claim span.  Use of hyperlink structure in Wikipedia or performing entity linking to establish hyperlinked mentions, similar to our approach here, has been previously explored in multi-hop open domain question answering~\citep{asai2020learning,nie-etal-2019-revealing}.  
Mutations with a ``{\sc Del}'' instead of an evidence span, and an ``{\sc Ins}'' instead of a claim span, are treated as deletions and insertions of claim and evidence spans respectively.

\subsection{NatOp Assignment}
\label{ssec:entailent}

As shown in Figure~\ref{fig:annotation}, the NatOp assignment step produces a sequence of NatOps, one for each mutation. Here, the search space becomes exponentially large, i.e.\ $6^n$ possible NatOp sequences for $n$ mutations. First, we assign NatOps to individual mutations  relying on hand-crafted rules and external resources, without considering the other mutations in the sequence 
(\S~\ref{subsec:individual}). With this partially filled NatOp sequence, we perform two filtering steps to further reduce the search space. We describe these steps below: one using veracity label information from training data in FEVER \cite{thorne-etal-2018-fever} and another using some additional manual annotation information from annotation logs of FEVER (\S~\ref{filteringAnnot}). 


\subsubsection{Initial Assignment}
\label{subsec:individual}

The initial assignment of NatOps considers each mutation in the sequence in isolation. 
Here,  mutations which fully match lexically are assigned with  the equivalence NatOp~($\equiv$), like the mutations 1, 4 and 5  in Figure~\ref{fig:annotation}. Similarly, mutations where the claim or evidence span has an extra negation word but lexically match otherwise, are assigned the negation NatOp~($\curlywedge$). Further, insertions and deletions, i.e.\ mutations with {\sc Ins} and {\sc Del} respectively (\S \ref{ssec:chunkALign}), containing negation words are also assigned the negation NatOp. To obtain these words, we identify a set of common negation words from the list of stop words in \newcite{spacy}, and combine them with the list of negative sentiment polarity words from  \newcite{10.1145/1014052.1014073SentimentPolarity}. 
Remaining cases of insertions (deletions) are treated as making the existing claim more specific (general), and hence assigned the forward (reverse) entailment NatOp, like mutation 3 in Figure \ref{fig:annotation}. Furthermore, as every paraphrase pair present in Paraphrase Database~\cite[PPDB][]{ganitkevitch-etal-2013-ppdb,pavlick-etal-2015-ppdb2} is marked with an entailment relation, we identify mutations which 
are present in it as paraphrases and assign the corresponding NatOp.



In several cases, the NatOp information need not be readily available 
at the span level. Here, we retain the word-level alignments from the aligner and perform lexical level NatOp assignment with the help of Wordnet~\cite{10.1145/219717.219748wordnet} and Wikidata~\cite{10.1145/2629489Wikidata}. We follow \newcite[Chapter 6]{maccartney2009natural} for NatOp assignment of open-class terms using Wordnet.



\begin{figure}[ht]
    \includegraphics[width=0.35\textwidth,trim={0 0.7cm 1cm 0cm}]{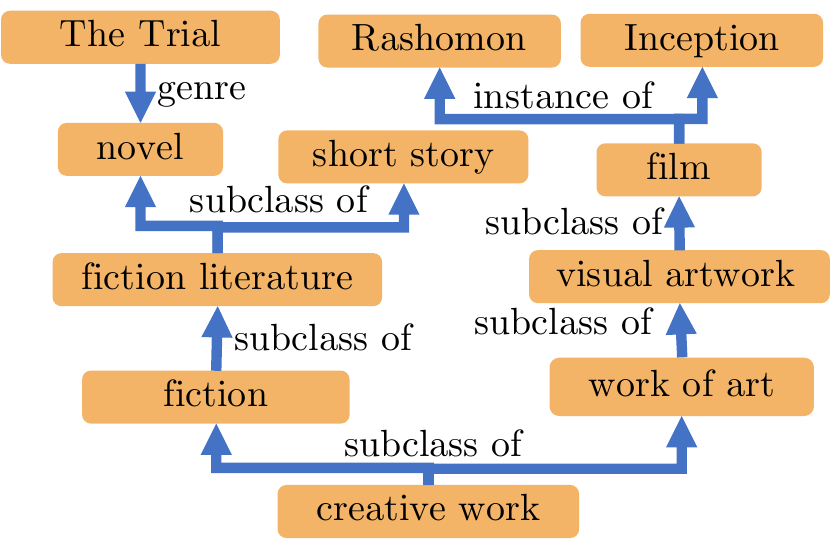}
    \caption{Entities and their relations in Wikidata. }
    \label{fig:wikidataKBs}
\end{figure}

Additionally, we define rules to assign a NatOp for named entities using Wikidata. Here, aliases of an entity are marked with an equivalence NatOp~($\equiv$), as shown in third triple in Figure~\ref{fig:working}. Further, we manually assign NatOps to the 500 most frequently occurring Wikidata relations in the aligned training data.  
For instance, as shown in Figure~\ref{fig:wikidataKBs},
 the entities `The Trial' and `novel' have the relation `genre'.
A claim span containing `The Trial', when substituted with an evidence span containing `novel', would result in a generalisation
of the claim, and hence will be assigned the reverse entailment NatOp ($\sqsupseteq$). A substitution in the reverse direction would be assigned a forward entailment NatOp ($\sqsubseteq$), indicating specialisation.



The KB relations we annotated 
occur between the entities linked in Wikidata, and they do not capture hierarchical multihop relations between the entities in the KB. We create such a hierarchy by combining the ``instance of'', ``part of'', and ``subclass of'' relations in Wikidata. 
Thus, a pair of entities connected via a directed path of length $k \leq 3 $, such as ``Work of art'' and ``Rashomon'' in Figure~\ref{fig:wikidataKBs}, is considered to have a parent-child relation, and assigned the forward or reverse entailment NatOp, depending on which span appears in the claim and the evidence.
Similarly, two entities, e.g.\ ``Rashomon'' and ``Inception'', are considered to be siblings if they have a common parent, and are assigned the alternation NatOp ($\updownharpoons$). 
However, two connected entities that do not satisfy the aforementioned distance criterion, e.g.\ ``novel'' and ``Rashomon'', are assigned with the independence NatOp ($\hash$), signifying they are unrelated.


\begin{table}[ht]
\centering
\begin{tabular}{|l|l|l|l|}
\hline
\multicolumn{1}{|c|}{Transformation}  & S & R & N \\ \hline

\begin{tabular}[c]{@{}l@{}}substitute with similar info.\end{tabular}     & $\sqsubseteq$   & $\updownharpoons$  &   $\sqsupseteq$  \\ \hline
\begin{tabular}[c]{@{}l@{}}substitute with dissimilar info.\end{tabular} &   $\sqsubseteq$ & $\updownharpoons$   &  $\hash$   \\ \hline
paraphrasing                                                                      &   $\equiv$   &  $\updownharpoons$  &  $\hash$  \\ \hline
negation                                                                          & $\curlywedge$  &  $\curlywedge$ & $\curlywedge$   \\ \hline
transform to specific                                                             & $\sqsubseteq$   & $\sqsubseteq$    &  $\sqsubseteq$   \\ \hline
transform to general                                                              & $\sqsupseteq$   & $\sqsupseteq$   & $\sqsupseteq$   \\ \hline
\end{tabular}
\caption{NatOp assignment based on transformations and veracity label information.}
\label{tab:correction}
\end{table}

\subsubsection{Filtering the Search Space}
\label{filteringAnnot}
While in Section \ref{subsec:individual} we assigned a NatOp to each mutation in isolation, there can still be unfilled NatOps. For instance, the unfilled NatOp in the second mutation 
of Figure~\ref{fig:annotation} leads to six possible  NatOp sequences as candidates, one per available NatOp. 
Recall that these NatOp sequences act as a transition sequence in the DFA (\S~\ref{dfaDecision}).  Thus we make use of the partially filled NatOp sequence and the veracity label from the training data to filter out NatOp sequences that do not terminate at the same state as the veracity label according to the DFA. The instance in Figure~\ref{fig:annotation} has the {\sc Support} label, and among the six possible candidate sequences only two terminate in this label. Hence, we retain those two sequences.





For the final filtering step we use the additional manual annotation that was produced during the construction of the claims in FEVER. There, the annotators constructed each claim by manipulating a factoid extracted from Wikipedia using one of the six transformations listed in Table~\ref{tab:correction}. Our proofs can be viewed as an attempt at reconstructing the factoid from a claim in multiple mutations, whereas these transformations can be considered claim-level mutations that transition directly from the last step (reconstructed factoid) in the proof to the first step (claim). This factoid is treated as the corrected claim in \citet{thorne2021factual} who released this annotation. 
For each veracity label we define the mapping of each transformation to a NatOp, as described in Table~\ref{tab:correction}. The assumption is that if a transformation has resulted in a particular veracity label, then the corresponding NatOp is likely to occur in the proof. To identify the mutation to assign it, we obtain the text portions in the claim manipulated by the annotators to construct it, by comparing the claim and the original Wikipedia factoid. In the example of Figure~\ref{fig:annotation}, this transformed text span happens to be part of the second mutation, and as per Table~\ref{tab:correction} forward entailment is the corresponding NatOp given the veracity label,  resulting in the selection of the first NatOp sequence. In rare occasions (2.55\% claims in FEVER), we manually performed NatOp assignment, as the filtering steps led to zero candidates in those cases. As the heuristic annotation requires manual effort,
we explore how it can be obtained using a supervised classifier (see \S \ref{dataHeur}).

\section{Experimental Methodology}

\subsection{Data}
\label{data}

\modelName~is trained using heuristically annotated proofs~(\S \ref{sec:explanation_training}) obtained from FEVER~\cite{thorne-etal-2018-fever}, which has a train-test-development split of 
145,449, 19,998, and 19,998 claims respectively. Further, the heuristic proof annotation involves the use of additional information from the manual annotation logs of FEVER, recently released by \newcite{thorne2021factual}. Finally, claims with the label {\sc Not Enough Info (NEI)} require retrieved evidence for obtaining their proofs for training, as no ground truth evidence exists for such cases. Here, we use the same retriever that would be used during the prediction time as well.





In addition to FEVER, we train and evaluate \modelName~on two other related datasets. First, we use Symmetric FEVER~\cite{schuster-etal-2019-towards}, a dataset designed to assess the robustness of fact verification systems against the claim-only bias present in FEVER. The dataset consists of 1,420 counterfactual instances, split into development and test sets of 708 and 712 instances respectively. Here, we heuristically generate the ground truth proofs for the dataset's development data and use it to fine tune \modelName, before evaluating it on the dataset's test data. Similarly, we also evaluate \modelName~on the FEVER 2.0 adversarial examples \cite{thorne-etal-2019-fever2}. Specifically, we use the same evaluation subset of 766 claims that was used by \newcite{schuster-etal-2021-get}. To fine-tune  \modelName~on this dataset, we generate the ground truth proofs for 2,100 additional adversarial claims, 
separate from the
evaluation set, which were curated by the organisers and participants of the FEVER 2.0 shared task. 


Finally, we also use the manual annotation logs of FEVER \cite{thorne2021factual} to obtain rationales for 
claims in the development data. In particular, we obtain the rationale for a claim  by extracting from its corresponding Wikipedia factoid the words  which were removed by the annotators during its creation. If these words are part of an evidence sentence, then they become the rationale for veracity label of the claim given the evidence. Further, we require that the words extracted as rationale 
form a contiguous phrase. We identified 300 claims which satisfy all these criteria.

\subsection{Evaluation Metrics} 

The evaluation metrics for FEVER are label accuracy (LA, i.e.\ veracity accuracy) and FEVER Score~\cite{thorne-etal-2018-fact}, 
which rewards only those predictions which are accompanied by  at least one correct set of evidence sentences. We report  mean LA and standard deviation for 
experiments with Symmetric FEVER,  where we use its development data for training and train with five random initialisations 
due to its limited size.


We further introduce a new evaluation metric, to assess model robustness,
called Stability Error Rate (SER). Neural models, especially with a retriever component,
have shown to be vulnerable to model overstability~\cite{jia-liang-2017-adversarial}.
Overstability is the inability of a model to distinguish superfluous information which merely has lexical similarity with the input, from the information truly relevant to arrive at the correct decision. 
In the context of fact verification,
it is expected that an ideal model should always predict {\sc Not Enough Info}, whenever it lacks sufficient evidence to make a decision otherwise. Further, it should arrive at a  {\sc Refute} or {\sc Support}  decision only when the model possesses sufficient evidence to do so, and any additional evidence should not alter its decision. 
 To assess the model overstability in fact verification, we define SER 
as the percentage of claims where additional evidence alters the {\sc Support} or {\sc Refute} decision of a model.

\subsection{Baseline Systems}

\paragraph{KGAT~\cite{liu-etal-2020-fineKGAT}}  uses a graph attention network, 
where each evidence sentence, concatenated with the claim, forms a node in the graph. We use their best configuration, where the node representations are initialised using RoBERTA~(Large). The relative importance of each node is computed with node kernels, and information propagation is performed using edge kernels. They also propose a new evidence sentence retriever, a BERT model trained with a pairwise ranking loss, though they rely on past work for document retrieval \cite{hanselowski-etal-2018-ukp}.



\paragraph{CorefBERT~\cite{ye-etal-2020-coreferential}}  follows KGAT and differs only in terms of the LM used for the node initialisation. Here, they further pretrain the LM on a task that involves prediction of referents of a masked mention to capture co-referential relations in context. We use CorefRoBERTA, their best-performing configuration. 


\paragraph{DominikS \newcite{stammbach-2021-evidence}}
focuses primarily on sentence-level evidence retrieval,  scoring individual tokens from a given Wikipedia document, and then selecting the highest scoring sentences by averaging token scores. It uses a fine-tuned document level BigBird model \cite{bigbird} for this purpose. 
For claim verification it
uses a DeBERTa \cite{he2021deberta} based classifier.

\subsection{\modelName: Implementation Details}

We follow most previous works on FEVER which model the task in 
three steps, namely 
 document retrieval, retrieval of evidence sentences from them, and finally veracity prediction based on 
 the evidence. 
 \modelName's novelty lies in the proof generation in the third step. Hence, for better comparability, we follow two popular, well-performing retrieval approaches, \newcite{liu-etal-2020-fineKGAT} and \newcite{stammbach-2021-evidence}.  
 \newcite{liu-etal-2020-fineKGAT}'s sentence retriever, also used in \newcite{ye-etal-2020-coreferential}, is a sentence level pairwise ranking model, whereas that of \newcite{stammbach-2021-evidence} is a document level token score aggregation model. \modelName's configuration which uses the former is our default configuration, referred to as \modelName,  and the configuration using the  latter will  henceforth be referred to as \modelName-SB. We retrieve five sentences for each claim as required in the FEVER evaluation.



For the proof generator, we use the pretrained BART~(Large) model~\cite{lewis-etal-2020-bart} and fine tune it using the heuristically annotated data from Section \ref{sec:explanation_training}. During prediction, the search spaces for the claim and evidence are populated using two separate tries.  We add all possible subsequences of the claim and evidence, each with one to seven words, into the respective tries. 
The default configuration takes the concatenation of a claim and all the retrieved evidence together as a single input, separated by a delimiter. 

We consider three additional configurations
which differ in the way the retrieved evidence is handled. In \modelName-MV, a claim is concatenated with 
one evidence sentence at a time;
this produces five proofs and five decisions per claim, and the final label is decided based on majority voting (MV).
Both  \modelName-A and -AR are designed to 
restrict the  proof generator's flexibility in inferring the textual spans in the mutations,
and thus assess the gains obtained by allowing it in \modelName. \modelName-A (aligned) considers during prediction only the subsequences from each evidence sentence aligned with the claim using word-level alignment, which are then concatenated with the claim as its input during training and prediction. 
Thus, the evidence search space 
becomes narrower, as the unaligned portions in the evidence are not considered.
\modelName-AR (aligned-restricted) further restricts the search space of both the claim and evidence,
by predetermining the number of mutations, the claim spans in these mutations and five candidate evidence spans for each mutation (one per evidence sentence). It obtains this information using the chunker and aligner used in the heuristic annotation (\S \ref{sec:explanation_training}). 

\subsection{Heuristic Annotation Using Kepler  }
\label{dataHeur}

To reduce the reliance on
manual annotation
from \newcite{thorne2021factual} during the annotation in Section \ref{sec:explanation_training}, 
we experiment with replacing the ground truth transformations with predicted ones using a classifier. We use KEPLER~\cite{TACL2447Kepler}, a RoBERTA-based pretrained LM enhanced 
with KB relations and entity pairs from WikiData for the classification. KEPLER covers 97.5~\% of the entities present in FEVER.  We first train it  with the FEVER training dataset for the fact verification task. Then we fine-tune it for the six-class classification task of predicting the  transformations, given a claim, evidence sentence and veracity label as input from the FEVER training data. We train it with varying training dataset sizes ranging from 1.24\% (1,800; 300 per class) to 41.24\% (60,000; 10,000 per class) of the FEVER training data. We consider two configurations: 
\modelName-K which uses gold data to identify the transformed span for applying the predicted transformation, and 
\modelName-K-NoS which instead 
only ensures that the predicted transformation occurs at least once in the final NatOp sequence.

\section{Results}

\begin{table}[!t]
\centering
\begin{tabular}{|crrrr|}
\hline
\multicolumn{1}{|c|}{\multirow{2}{*}{System}} & \multicolumn{2}{c|}{Dev}                                                                                         & \multicolumn{2}{c|}{Test}                                                                                         \\ \cline{2-5} 
\multicolumn{1}{|c|}{}                        & \multicolumn{1}{c|}{LA}             & \multicolumn{1}{c|}{\begin{tabular}[c]{@{}l@{}}Fever\\ Score\end{tabular}} & \multicolumn{1}{c|}{LA}             & \multicolumn{1}{c|}{\begin{tabular}[c]{@{}l@{}}Fever\\ Score \end{tabular}} \\ \hline
\hline
\multicolumn{5}{|c|}{
Using Retriever from \newcite{liu-etal-2020-fineKGAT}}
\\\hline

\multicolumn{1}{|c|}{\modelName}              & \multicolumn{1}{r|}{\textbf{80.23}} & \multicolumn{1}{r|}{\textbf{78.17 }}                                       & \multicolumn{1}{r|}{\textbf{79.25}} & \textbf{74.37 }                                                             \\ \hline
\multicolumn{1}{|c|}{\modelName-MV}           & \multicolumn{1}{r|}{78.71}          & \multicolumn{1}{r|}{74.62}                                                 & \multicolumn{1}{r|}{74.18}          & 70.09                                                                       \\ \hline
\multicolumn{1}{|c|}{\modelName-A}            & \multicolumn{1}{r|}{79.83}          & \multicolumn{1}{r|}{76.33}                                                 & \multicolumn{1}{r|}{77.16}          & 72.47                                                                       \\ \hline
\multicolumn{1}{|c|}{\modelName-AR}           & \multicolumn{1}{r|}{77.42}          & \multicolumn{1}{r|}{75.27}                                                 & \multicolumn{1}{r|}{-}              & -                                                                           \\ \hline
\multicolumn{1}{|c|}{KGAT}                    & \multicolumn{1}{r|}{78.29}          & \multicolumn{1}{r|}{76.11}                                                 & \multicolumn{1}{r|}{74.07}          & 70.38                                                                       \\ \hline
\multicolumn{1}{|c|}{CorefBERT}               & \multicolumn{1}{r|}{79.12}          & \multicolumn{1}{r|}{77.46}                                                 & \multicolumn{1}{r|}{75.96}          & 72.30                                                                       \\ \hline
\multicolumn{5}{|c|}{
Using Retriever from \newcite{stammbach-2021-evidence}}
\\ \hline
\hline
\multicolumn{1}{|l|}{\modelName-SB}           & \multicolumn{1}{r|}{\textbf{80.74}}               & \multicolumn{1}{r|}{\textbf{79.07}}                                                      & \multicolumn{1}{r|}{\textbf{79.47}}               &     \textbf{76.82}                                                                        \\ \hline
\multicolumn{1}{|l|}{DominikS}                & \multicolumn{1}{r|}{80.59}               & \multicolumn{1}{r|}{78.37}                                                      & \multicolumn{1}{r|}{79.16}               &     76.78                                                                        \\ \hline
\end{tabular}
\caption{Fact verification results 
on FEVER.}
\label{tab:mainPerformance}
\end{table}

\subsection{Fact Verification}
\label{ssec:performance}

Table~\ref{tab:mainPerformance} reports the fact verification results for \modelName~and the baselines. Overall,  \modelName-SB, our configuration using \newcite{stammbach-2021-evidence}'s retriever, is the best performing model in our experiments. \modelName-SB, which outperforms \newcite{stammbach-2021-evidence} itself, is currently the highest scoring model in terms of label accuracy in the FEVER leaderboard. It also is the second best model in terms of FEVER Score, second only to the currently unpublished model titled ``mitchell.dehaven'', in the leaderboard.

\modelName, our default configuration using the retriever from \newcite{liu-etal-2020-fineKGAT}, differs from \modelName-SB only in terms of the retriever they use. \modelName~is the best performing model among all the baselines and other \modelName~configurations (-MV, -A and -AR) that use \newcite{liu-etal-2020-fineKGAT}'s retriever. 
%
As compared to \modelName-MV, \modelName's gains come primarily from its ability to handle multiple evidence sentences 
together, 
as opposed to handling each separately and then aggregating the predictions. 
9.8\% (1,960) of the claims in the FEVER development set require multiple evidence sentences for verification. While \modelName-MV predicts 60.1\% of these instances  correctly, \modelName~correctly predicts 67.45\% of these. Further, around 80.73\% (of 18,038) of the single evidence instances are correctly predicted by \modelName-MV, in comparison to 81.62\% instances for \modelName. 
Allowing the proof generator to infer the mutations dynamically, instead of having them predefined, benefits the overall performance of the model. The increasingly restricted variants with narrower search spaces, i.e.\ \modelName-A~and \modelName-AR, lead to decreasing performances as shown in Table \ref{tab:mainPerformance}. 
\modelName-AR, the most restricted version, performs worse than all the other models. 

\begin{table}[!ht]
\centering
\begin{tabular}{|c|c|c|c|}
\hline
\multicolumn{2}{|c|}{KEPLER} & \multicolumn{2}{c|}{ProoFVer} \\ \hline
\begin{tabular}[c]{@{}c@{}}Training \\ Data Size\end{tabular} & \begin{tabular}[c]{@{}c@{}}Classifier \\ Accuracy\end{tabular} & \begin{tabular}[c]{@{}c@{}}-K-NoS\\ (LA)\end{tabular} & \begin{tabular}[c]{@{}c@{}}-K\\ (LA)\end{tabular} \\ \hline

1,800   & 69.07  &  64.65 & 66.73            \\ \hline
6,000                                                         & 74.02  &  68.86  & 72.41             \\ \hline
18,000  & 79.67  & 74.25 & 76.23              \\ \hline
30,000                                                & 80.61  & 75.39 & 77.76              \\ \hline
45,000            & 82.76  & 77.62  & 78.84              \\ \hline
60,000                        & 84.85 & 78.61  & 79.67                \\ \hline
\end{tabular}

\caption{LA of \modelName-K and -NoS using predictions from KEPLER. Training data size used for
KEPLER and its classifier accuracy is also provided.}
\label{tab:kepler}
\end{table}

\begin{table*}[!ht]
\centering
\begin{tabular}{|c|c|c|c|c|c|c|}
\hline
\multirow{3}{*}{Model} & \multicolumn{6}{c|}{Dataset}                                                                        \\ \cline{2-7} 
                       & \multicolumn{3}{c|}{FEVER-DEV}                   & \multicolumn{3}{c|}{Symmetric FEVER}             \\ \cline{2-7} 
                       & Original       & FT      & FT+L2 & Original       & FT      & FT+L2 \\ \hline
\modelName             & \textbf{89.07$\pm$0.3} & \textbf{86.41$\pm$0.8} & \textbf{87.95$\pm$1.0$^*$} & \textbf{81.70$\pm$0.4} & \textbf{85.88$\pm$1.3$^\#$}          & \textbf{83.37$\pm$1.3$^{\#*}$} \\ \hline
KGAT              & 86.02$\pm$0.2         & 76.67$\pm$0.3       & 79.93$\pm$0.9$^*$          & 65.73$\pm$0.3          & 84.94$\pm$1.1$^\#$          & 73.34$\pm$1.5$^{\#*}$          \\ \hline
CorefBERT              & 88.26$\pm$0.4          & 78.79$\pm$0.2          & 84.22$\pm$1.5$^*$          & 68.49$\pm$0.6          & 85.45$\pm$0.2$^\#$ & 77.37$\pm$0.5$^{\#*}$          \\ \hline
\end{tabular}
\caption{Label accuracy of models on FEVER-development(DEV) and Symmetric FEVER with and without fine tuning.  All results marked with $*$ and $\#$ are statistically significant (unpaired t-test) with $p < 0.05$ against their FT and Original variants respectively. FEVER-DEV predictions are using gold standard evidence.  }
\label{tab:robustness}
\end{table*}

\paragraph{Impact of additional manual annotation}%

Since the final filtering step in NatOp assignment (\S \ref{filteringAnnot}) requires additional manual annotation, we experimented with a proof set obtained without this step. Here, we arbitrarily select a NatOp sequence from the candidates remaining after the veracity label based filtering. The latter
reduced the search space to just two possible NatOp sequences in 93.59\% of the claims. However, training \modelName~with these proofs resulted in a LA of 58.29\% on the FEVER development set. In comparison, \modelName-K-NoS achieves a LA of 64.65\%, even when using predictions from a KEPLER configuration trained on as little as 1,800 instances. Table \ref{tab:kepler} shows the LA for  \modelName-K-NoS and \modelName-K when using KEPLER predictions,  with varying training data sizes for KEPLER; the largest KEPLER configuration is trained  on only 41.24\%  of claims in FEVER. 
Using this amount of training data, \modelName-K and \modelName-K-NoS achieve a LA of 79.67\% and 78.61\% respectively. Here, \modelName-K outperforms all the baseline models, including CorefBert which 
also uses additional annotation for pretraining.

\subsection{Robustness}
\label{ssec:robustness}


\paragraph{Symmetric FEVER} As shown in Table~\ref{tab:robustness},  \modelName\ shows better robustness with a mean accuracy of 81.70\% on the Symmetric FEVER test dataset, an improvement of 13.21\%  over CorefBERT, the next best model. 
All models improve their accuracy and are comparable on the test set when we fine-tune them on its development set. However, this results in more than 9\% reduction on the original FEVER-DEV data for both the classifier based models, KGAT and CorefBERT. This catastrophic forgetting~\cite{FRENCH1999128catastrophic} occurs primarily due to the  shift in label distribution during fine tuning, as Symmetric FEVER contains only claims with  {\sc Support} and {\sc Refute} labels. \modelName\ accuracy drops by only less than 3\%, as it is trained with a seq2seq objective. To mitigate the effect of catastrophic forgetting, we apply L2 regularisation~\cite{thorne-vlachos-2021-elastic} which improves all models on the FEVER development set. Nevertheless, \modelName~ 
has the highest accuracy on both FEVER and Symmetric FEVER among the competing models after regularisation.

 \begin{figure*}[!ht]
    \centering
    \includegraphics[width=0.9\textwidth,page=15,trim={0 3.1cm 1cm 0cm}]{images/figures.pdf}
    \caption{Human rationale extraction for predicted proofs from \modelName. The claim and evidence spans are enclosed within `\{ \}' and `[ ]' respectively, with numbered superscripts showing the correspondence between the spans. The predicted rationales are underlined and the portions matching with the human rationales are highlighted. } 
    \label{fig:rationale}
\end{figure*}

\paragraph{Generalising to FEVER 2.0} 
\modelName~when evaluated on FEVER 2.0 adversarial data, reports a LA of 82.79\%, outperforming the previously best reported LA of 82.51\% by \newcite{schuster-etal-2021-get}. \modelName, after training on FEVER, is further fine tuned (with L2 regularization) on
heuristically generated proofs from the data contributed by the participants of the FEVER 2.0 shared task (disjoint from the evaluation set), and the proofs generated from the FEVER Symmetric data. On the other hand,
 \newcite{schuster-etal-2021-get} was trained on the VitaminC training data. When they further fine tune their default model with FEVER, their performance drops to 80.94\%.



\paragraph{Stability Error Rate (SER):}  SER 
quantifies the rate of instances where a system alters its decision due additional evidence in the input, passed on by the retriever component. 
KGAT, CorefBERT, and DominikS have a SER of 12.35\%, 10.27\%, 9.36 \% respectively. \modelName~has an SER of only 6.21\%, which is further reduced to 5.73 \% for \modelName-SB. The SER results confirm that the baselines change their predictions from {\sc Support} or {\sc Refute} after providing them with additional information more often than \modelName. 

\subsection{\modelName~proofs as explanations}
\label{ssec:explanationResults}



\subsubsection{Rationale Extraction}
Rationales extracted based on attention are often used as means to highlight the reasoning involved in the decision making process of various models~\cite{deyoung-etal-2020-eraser}. For this evaluation, we compare using token-level F-score of the predicted rationales with human-provided rationales 
for 300 claims 
from the FEVER development data, as elaborated in Section \ref{data}. We ensure that all the systems 
are provided with the same set of evidence sentences, 
and consider only those words from the evidence as rationales which do not occur in the claim. For \modelName, we additionally remove evidence spans which are part of mutations with an equivalence NatOp. 
For KGAT and CorefBERT, we obtain the rationales by sorting the eligible words in descending order of their attention scores, and for each instance we find the set of words with the highest token overlap F-score with the rationale. Here, we consider the words in the top 1\% of attention scores, and also those ranging from 5\% to 50\% of the words in step sizes of 5\%. 
We find that \modelName~achieves a token level F-score of 93.28, compared to 87.61 and 86.42, the best F-Scores for CorefBERT and KGAT.  Figure \ref{fig:rationale} shows the rationales for 3 instances extracted from \modelName, one for each label. 
All the three proofs result in correct decisions. While for the first two claims there is a perfect overlap with the human rationale, the third claim in Figure \ref{fig:rationale} has some extraneous information in the predicted proof. 








\begin{table}[!ht]
\begin{tabular}{|l|l|}
\hline
$\equiv$ & Equivalent Spans                                \\ \hline
$\updownharpoons$ & Evidence span contradicts the claim span        \\ \hline
$\sqsubseteq$ & Claim span follows from evidence span           \\ \hline
$\sqsubseteq$ & (Insert) New information from evidence \\ \hline
$\sqsupseteq$ & Incomplete Evidence                             \\ \hline
$\curlywedge$ & Evidence span refutes claim span                \\ \hline
$\curlywedge$ & Claim span negated (Deletion)                   \\ \hline
$\hash$  & Unrelated claim span and evidence span          \\ \hline
$\hash$ & No related evidence found (Deletion)            \\ \hline
\end{tabular}
\caption{NatOPs and the corresponding paraphrases}
\label{tab:parpahrase}
\end{table}

\begin{figure*}[!ht]
    \centering
    \includegraphics[width=0.9\textwidth,page=16,trim={0 4.2cm 1cm 0cm}]{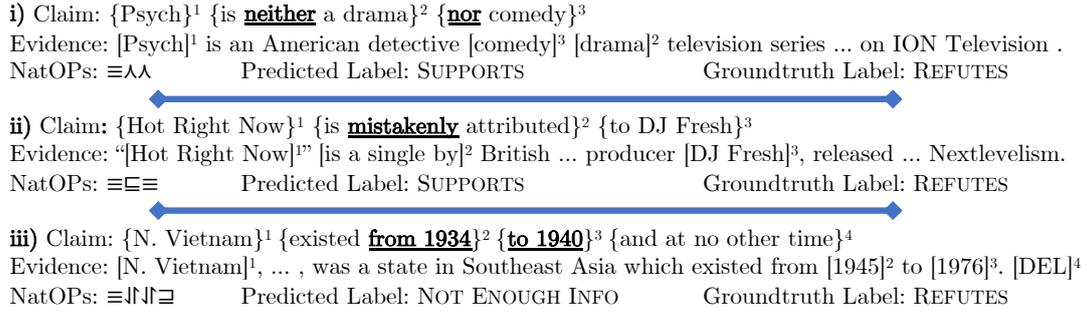}
    \caption{Cases of incorrect proof generation from \modelName. The claim and evidence spans are enclosed within `\{ \}' and `[ ]' respectively, with numbered superscripts showing the correspondence between the spans. } 
    \label{fig:mistakes}
\end{figure*}

\subsubsection{Human evaluation} We use forward prediction~\cite{doshivelez2017rigorous} here, where humans are asked to predict the system output based on the explanations. 
For assessing \modelName,
we provide the claim, the proof as the explanation, and those evidence sentences from which the evidence spans in the proof were extracted. 
Since we are interested in evaluating the applicability of our proofs as  natural language explanations, 
 we ensure that none of our subjects are aware of the deterministic nature of determining the label from natural logic proofs. Moreover, we replaced the NatOps in the proof with plain English phrases for better comprehension by the subjects, as shown in Table \ref{tab:parpahrase}. 
As the baseline setup for comparison, we provide the claim 
with all five retrieved evidence sentences. 

We form a set of 24 different claims, 12 each from \modelName~and baseline, and 3 individual subjects independently annotate the same set. Finally, we altogether obtain annotations for 5 sets, resulting in 60 claims, 120 explanations and a total of 360 annotations from 15 subjects.\footnote{Although 19 subjects volunteered, one of them annotated a set which did not receive any other annotations. In another set, two of them  had prior knowledge in natural logic leading to disqualification of all these 3 annotations from the set.} 
For all 60 claims, \modelName, CorefBERT and KGAT predicted the same labels, though not necessarily the correct ones (the subjects were not aware of this). 
 All the subjects were pursuing a PhD or postdocs in fields related to computer science and computational linguistics, or industry researchers/data scientists.  

With \modelName's proofs, subjects are able to predict the model decisions correctly in 81.67\% of the cases as against 69.44\% of the cases with only the evidence. In both setups, subjects were often confused on instances with a {\sc Not Enough Info} label, and the forward predictions were comparable, with 
66.67\%~(\modelName) and 65\% (baseline). In many such cases, subjects subconsciously filled in their own world knowledge which is not found in the evidence to arrive at a {\sc Support} or {\sc Refute} decision. Further, for instances with both {\sc Refute} and {\sc Support} labels, subjects correctly predicted \modelName's decisions 86.67\%  and 91.67\% times respectively, against only 70\% and 73.33\% for the baseline. The inter-annotator agreement for \modelName's explanations is 0.7074 in Fleiss $\kappa$ \cite{fleiss1971measuring}, and 0.6612 for the baseline.





\section{Limitations}

Figure \ref{fig:mistakes} shows three instances of incorrect proofs from \modelName, which highlight some of the well known limitations in natural logic \cite{karttunen2015natural,maccartney2009natural}.  In Figure \ref{fig:mistakes}.i, the claim uses two negation words, ``neither'' and ``nor'', both of which appearing in different spans and leading to prediction of two negation NatOps. However, this NatOp sequence nullifies the effect of the negation NatOp and predicts {\sc Support} instead of {\sc Refute}. Similarly, in Figure~\ref{fig:mistakes}.ii the adverb ``mistakenly'' negates semantics of the verb. However, its effect is not captured in the second mutation and \modelName~predicts the forward entailment NatOP, leading to the {\sc Support} label. Moreover, the NatOP sequence remains the same even if we remove the term ``mistakenly'' from the claim, demonstrating that the effect of the adverb is not captured by our model. Similar challenges involving adverbs and non-subsective adjectives \cite{pavlick-callison-burch-2016-called} when performing inference in natural logic has been reported in prior work \cite{angeli-manning-2014-naturalli}. 

In Figure \ref{fig:mistakes}.iii, the claim states a time period by mentioning its start and end years, which appear in two different claim spans. However, \modelName~does not capture the sense of the range implied by the spans containing ``from 1934'' and ``to 1940''. Instead, two similar 4-digit number patterns are extracted from the evidence and are directly compared to the claim spans resulting in two alternation NatOps, thereby predicting {\sc Not Enough Info}. Handling such 
range expressions is beyond the expressive power of the natural logic, and often  other logical forms are needed to perform such computations \cite{liang-etal-2013-learning}. Datasets like FEVEROUS~\citep{Aly2021FEVEROUSFE}, which considers semi-structured information present in tables, often require such explicit computations for which approaches purely based on natural logic are not sufficient.

Finally,  \modelName, due to its auto-regressive formulation, generates the corresponding evidence spans and NatOps for the claim spans sequentially from left to right. However, the steps in the natural logic based inference are not subject to any such specific ordering, and hence the order in which the NatOPs are generated is non deterministic by default \cite{angeli-manning-2014-naturalli}.  \modelName~benefits from the implicit knowledge encoded in the pretrained language models, specifically BART, which follows auto-regressive decoding.
Nevertheless, in the future we plan to experiment with alternative decoding approaches, including some of the recent developments in non-autoregressive conditional language models \cite{xu-carpuat-2021-editor} and 
transformer-based proof generators \cite{saha-etal-2021-multiprover}.



\section{Conclusion}

We presented \modelName, a natural logic-based proof system for fact verification. 
Currently, we report the best results in terms of label accuracy, and the second best results in FEVER Score in the FEVER leaderboard.
Moreover, \modelName~is more robust in handling superfluous information from the retriever, and handling counterfactual instances. 
Finally, \modelName 's proofs are faithful explanations  by construction, 
and improve the understanding of the decision making process of the models by humans.

\section*{Acknowledgements}

Amrith Krishna and Andreas Vlachos are supported by a Sponsored Research Agreement between Facebook (now Meta) and the University of Cambridge. Andreas Vlachos is additionally supported by the ERC grant AVeriTeC (GA 865958) and the EU H2020 grant MONITIO (GA 965576). The authors would like to thank Nicola De Cao for helpful conversations, and Dominik Stammbach for sharing the retriever data. Further, the authors would like to thank all the 19 subjects who volunteered to be part of the human evaluation, namely James Thorne, Michael Sejr Schlichtkrull, Ashim Gupta, Bishal Santra, Ashutosh Soni, Abhik Jana, Sudipa Mandal, Madhumita Mallick, Roshni Tayal, Sandipan Sikdar, Rishabh Kumar, Jasabanta Patro, Unni Krishnan, Anirban Santara, Jivnesh Sandhan, Soumya Sarkar, Devaraj Adiga, Aniruddha Roy and Muskan Garg.  Finally, we would like to thank the anonymous reviewers, action editor Mohit Bansal and the editor in chief Brian Roark for their valuable feedback.

\bibliography{tacl2018}
\bibliographystyle{acl_natbib}

\end{document}